# Deep Multi-task Prediction of Lung Cancer and Cancer-free Progression from Censored Heterogenous Clinical Imaging

Riqiang Gao[a*], Lingfeng Li[a], Yucheng Tang[b], Sanja L. Antic[c], Alexis B. Paulson[c], Yuankai Huo[a], Kim L. Sandler[c], Pierre P. Massion[c], Bennett A. Landman[a,b]

[a] Computer Science, Vanderbilt University, Nashville, TN, USA 37235
[b] Electrical Engineering, Vanderbilt University, Nashville, TN, USA 37235
[c] Vanderbilt University Medical Center, Nashville, TN, USA 37235

(*Corresponding Author: riqiang.gao@vanderbilt.edu)

## ABSTRACT

Annual low dose computed tomography (CT) lung screening is currently advised for individuals at high risk of lung cancer (e.g., heavy smokers between 55 and 80 years old). The recommended screening practice significantly reduces all-cause mortality, but the vast majority of screening results are negative for cancer. If patients at very low risk could be identified based on individualized, image-based biomarkers, the health care resources could be more efficiently allocated to higher risk patients and reduce overall exposure to ionizing radiation. In this work, we propose a multi-task (diagnosis and prognosis) deep convolutional neural network to improve the diagnostic accuracy over a baseline model while simultaneously estimating a personalized cancer-free progression time (CFPT). A novel Censored Regression Loss (CRL) is proposed to perform weakly supervised regression so that even single negative screening scans can provide small incremental value. Herein, we study 2287 scans from 1433 de-identified patients from the Vanderbilt Lung Screening Program (VLSP) and the Consortium for Molecular and Cellular Characterization of Screen-Detected Lesions (MCL) cohorts. Using five-fold cross-validation, we train a 3D attention-based network under two scenarios: (1) single-task learning with only classification, and (2) multi-task learning with both classification and regression. The single-task learning leads to a higher AUC compared with the Kaggle challenge winner pre-trained model (0.878 v. 0.856), and multi-task learning significantly improves the single-task one (AUC 0.895, $p<0.01$, McNemar test). In summary, the image-based predicted CFPT can be used in follow-up year lung cancer prediction and data assessment.

## 1. INTRODUCTION

Survival rates for patients with lung cancer are largely dependent on the stage of the disease at diagnosis and the feasible treatment plans [1]. If detected early, outcomes with modern therapy are quite positive. Current guidelines recommend annual low dose computed tomography (CT) lung screening for people with high risk of lung cancer (e.g., 55+ years old and a smoking history of 30+ pack years) (https://medlineplus.gov/lungcancer.html). A primary challenge with CT lung screening is that even in high risk screening populations, cancers are rare. For example, in the National Lung Screening Trail (NLST) [2] and Vanderbilt Lung Screening Program (VLSP) (https://www.vumc.org/radiology/lung), less than 5% of patients develop lung cancer. The National Lung Screening Trial (NLST) was used to design the current lung CT screening guidelines using risk factors like pack-years of smoking and age quit smoking. Compared with NLST criteria, the PLCOm2012 [3] brings significant improvement in sensitivity and positive predictive value by combining 11 different variables (e.g., age, chronic obstructive pulmonary disease (COPD), education). However, current risk models do not include prior CT screening in the risk level.

Much recent focus has been on detection of lung cancer in low dose screening CT. The prevalent methods include two steps: nodule detection and scan classification. Nodule detection finds the suspicious regions (region of interests, ROI) for lung cancer detection. Scan classification identifies a particular scan volume as cancer or non-cancer. Many researchers focus on the classification of the scan (e.g., [4][5][6][7]) or nodules (e.g., [8]). Ardila et al extended the concept of immediate classification of cancer to classification at a near-term (but future, e.g., 1 or 2 year) time point [5]. Meanwhile, other studies have investigated survival prediction with machine learning methods outside of low dose screening CT. For example, a lasso-regularized Cox model [9] is proposed for whole-slide images to predict survival of patients. The censored data, the measurement or observation is only partially known, are very common in medical research. Kaplan–Meier estimation [10] is a non-parametric statistic to estimate the survival function from lifetime data, which can take some censored data into account.

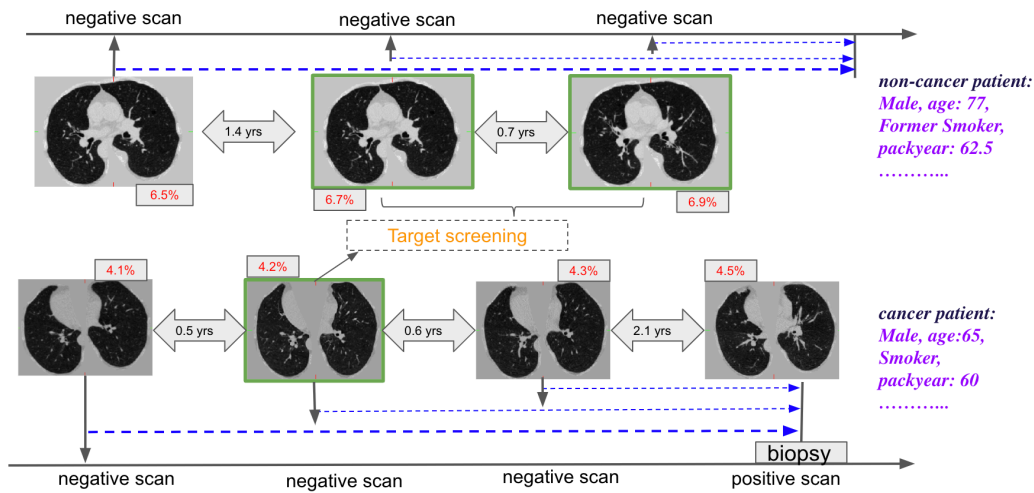

Figure 1. Vision for joint consideration of diagnosis and cancer-free progression time (CFPT) in cancer and non-cancer patients. The blue dashed lines represent the defined CFPT. The red number (%) in small rectangle is the risk score computed by the PLCOm2012 [3] model. Current scores suggest screening at every point since they indicate high risk (>= 2%). If CFPT were known, the green rectangles could have been avoided.

Herein, we build upon the deep learning classification results in lung screening CT, but also focus on prognosis and prediction of onset. We introduce the concept cancer-free progression time (CFPT) as the known duration between CT scan and the period for which the patient is known to not have had cancer, or the known period up to when a cancer diagnosis was made. By codifying this type of information, we can adapt the core cancer detection model into a multi-task consideration of detection along with CFPT. We hypothesize that if the CFPT can properly be predicted then this information could inform screening risk and guide follow-up studies. For example, patients at lower risk could avoid scans for longer period. Figure 1 illustrates both the problems with the risk models and our vision for how the CFPT information be use valuable.

Motivated by Kaplan–Meier estimation, we propose a new loss function in this paper, termed as Censored Regression Loss (CRL), to guide prediction of the occurrence of lung cancer. In contrast to prevalent regression loss functions (e.g., mean square error (MSE), mean absolute error (MAE), Huber loss), we define the loss function from a clinical perspective with censored data. We apply CRL in combination with the traditional cross entropy loss (CEL) to determine if CRL for CFPT prediction can significantly improve in cancer / non-cancer classification performance. We note that the CFPT metric itself also can be used as cancer prognosis / risk factor (e.g., following [5]).

## 2. METHODS

Clinical records of patients are usually censored data (especially right-censored), because patients have not completed their life experience, withdraw the study, or are lost to follow-up. The lack of a complete time record renders the CFPT time difficult to predict as a patient may either never be diagnosed with cancer or simply not be diagnosed cancer during the observed period. In this paper, we propose a new loss function to provide opportunities to learn new information in environments with varying level of risk within control populations.

**2.1 Defination of the cancer-free progression time (CFPT) and malignancy of CT scans**

**The CFPT of current scan.** For patients who develop cancer, the CFPT of a scan is defined as the time interval from scan to the biopsy time (as blue line shown in Figure 1). We set the last scan time as the biopsy time for those patients who are confirmed with lung cancer but the actual diagnose time is missing. For those patients who are not diagnosed with cancer, we assume that they will not have cancer within at least one year (e.g., the time until the next scheduled screening scan), and model the time interval as right censored after the last scan plus one year.

**Malignancy of CT scans.** In this paper, our target is to regress the cancer time of cancer patients, so we analyze the malignancy at scan-level rather than subject-level. We assume the diagnosis of lung cancer corresponds to the scan just before diagnose time. If the patient is diagnosed with lung cancer, the last scan not later than the diagnosis time and the scans after the diagnosis time are defined as malignant, and other scans are defined as negative of lung cancer.

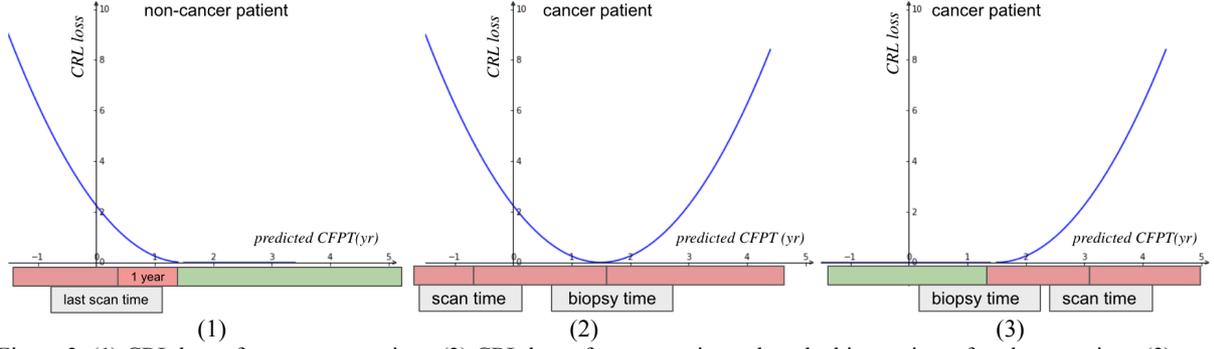

Figure 2. (1) CRL loss of non-cancer patient. (2) CRL loss of cancer patient when the biopsy time after the scan time. (3) CRL loss of cancer patient when the biopsy time before the scan time. The predicted CFPT located in the "red" part is with the loss from clinical perspective, while the "green" part is free of loss.

**2.2 Censored Regression Loss**

**Analyses of loss of predicting lung cancer time from the clinical perspective.**

As shown in Figure 2, we analyze the clinical loss (e.g., risk to envisioned clinical use from estimation errors) from three different conditions:

(1) For those patients not diagnosed with lung cancer, we assume that they will not have lung cancer within one year from the latest scan. Thus, there is no clinical loss if the predicted CFPT is larger than the defined CFPT (that is the last scan time plus one year for non-cancer patient).

(2) For those patients diagnosed with cancer and the scan time before the diagnose time. We regress the predicted CFPT to the defined CFPT.

(3) For those patients diagnosed with cancer and the scan time after the diagnose time. There is no clinical loss if we predict the CFPT smaller than the defined CFPT.

**The CRL formulation.**

According to the analyses of prediction loss from the proposed perspective, consider the following limitations: (1) actual CFPT is smaller than defined CFPT since a patient already developed cancer before having biopsy, (2) most patients without a cancer diagnosis will have more than 1-year cancer free follow up, and (3) a margin $\varepsilon$ in loss function contributes to feature discriminativeness as in [11][12][13]. Thus, we pose the following loss function equation:

$$L_r\left(t_i^{pred}, t_i^d\right) = \begin{cases} min\left(0, t_i^{pred} - t_i^d - \varepsilon\right)^2 & if\ p_i = 0 \\ \left(t_i^{pred} - t_i^d + \varepsilon\right)^2 & if\ p_i = 1\ and\ t_i^d > \varepsilon \\ max\left(0, t_i^{pred} - t_i^d + \varepsilon\right)^2 & if\ p_i = 1\ and\ t_i^d \leq \varepsilon \end{cases} \quad (1)$$

where $i$ is the scan-level index. The $t_i^{pred}$ is the predicted cancer time from the network, and $t_i^d$ is the diagnosis time defined in Section 2.2. The $p_i$ indicates if the scan $i$ comes from cancer patients or not, that is, $p_i = 1$ represents scan $i$ is the patient who is finally diagnosed as lung cancer. The $\varepsilon$ is a margin that is enforced between cancer patients and non-cancer patients.

**2.3 Joint training of regression and classification.**

The lung cancer detection of CT scans can be regarded as binary classification from a machine learning perspective. The most prevalent classification loss is Cross Entropy Loss (CEL), whose two-class version are defined as follows:

$$L_c(\hat{y}_i, y_i) = -y_i log(\hat{y}_i) - (1 - y_i)\log(1 - \hat{y}_i) \quad (2)$$

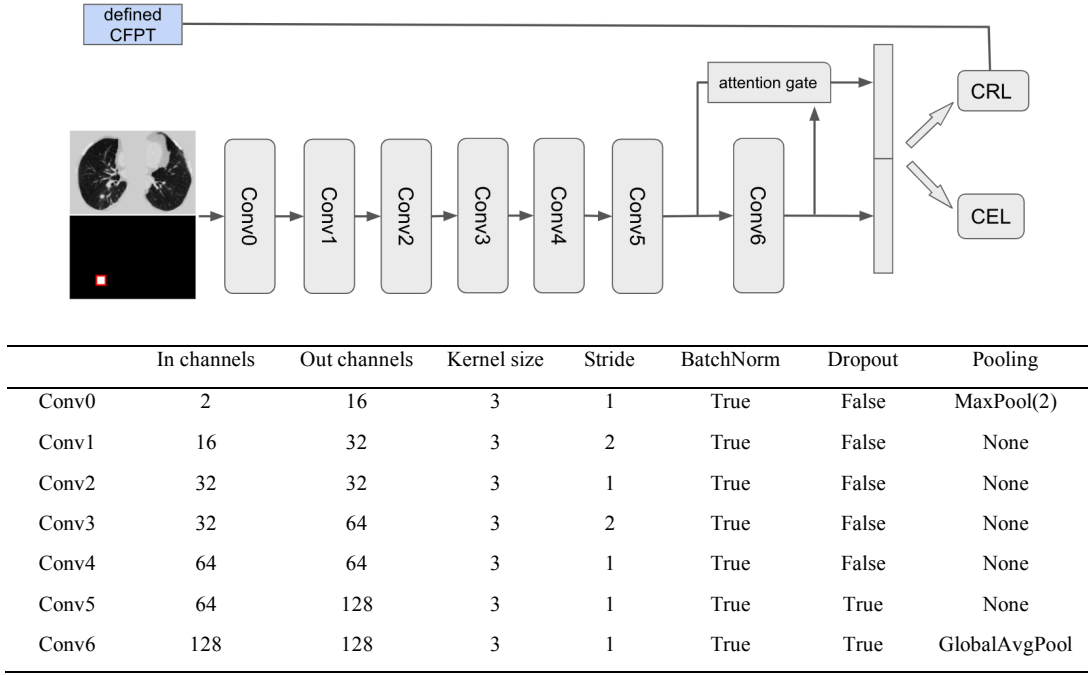

| | In channels | Out channels | Kernel size | Stride | BatchNorm | Dropout | Pooling |
|---|---|---|---|---|---|---|---|
| Conv0 | 2 | 16 | 3 | 1 | True | False | MaxPool(2) |
| Conv1 | 16 | 32 | 3 | 2 | True | False | None |
| Conv2 | 32 | 32 | 3 | 1 | True | False | None |
| Conv3 | 32 | 64 | 3 | 2 | True | False | None |
| Conv4 | 64 | 64 | 3 | 1 | True | False | None |
| Conv5 | 64 | 128 | 3 | 1 | True | True | None |
| Conv6 | 128 | 128 | 3 | 1 | True | True | GlobalAvgPool |

Figure 3. Network design. The upper part shows the network structure used in our paper. The input contains two parts: (1) two channels of 3-D CT volumes and its nodule mask, (2) diagnose time of the patient. Two losses (censored regression loss (CRL) for regressing the CFPT, cross entropy loss (CEL) for classifying the CT scan is cancer or not) are included in the training. The lower table is the details of layers which used in the network.

where $\hat{y}_i$ is the predicted probability that the scan $i$ belongs to lung cancer, and $y_i$ is the ground truth of scan $i$ (cancer or non-cancer).

The final loss function used in our training is:

$$L = \lambda \cdot L_r\left(t_i^{pred}, t_i^d\right) + L_c(\hat{y}_i, y_i) \tag{3}$$

where $\lambda$ is the hyperparameter that balances the regression loss and classification in the training.

## 3. DATA PREPROCESSING AND NETWORK

### 3.1 Image Preprocessing and Nodule Detection

**Image Preprocessing.** We follow the preprocessing pipeline of Liao et al. [4], whose method won the first place of Kaggle Challenge (Data Science Bowl 2017). The CT volumes are resampled to $1 \times 1 \times 1 mm^3$ isotropic resolution. The raw data are converted to Hounsfield Units (HU). The lung mask is created to rule out the impact of other tissues. Before feeding to the network, the raw data are clipped with the window [-1200, 600], and are normalized to [0, 255]. Afterwards, the data are converted from HU to UINT8 and the non-lung regions are padded with the value of 170.

**Nodule Detection.** Liao et al. [4] proposed a 3D version of the RPN along with modified U-Net for nodule detection. We use the open source code and pre-trained model (https://github.com/lfz/DSB2017) to detect nodules. The detection pipeline detects many nodule proposals and the non-maximum suppression (NMS) are used to rule out overlap.

### 3.2 Network Structure

As shown in Figure 3, the design of our network structure is inspired the work of [6], with small modifications to reduce GPU memory requirements. The details of the network are summarized in the bottom table of Figure 3. The raw CT scan is preprocessed and reshaped to $128 \times 128 \times 128$ before feeding to network. A 3D nodule mask with the same shape as reshaped data is generated with the output of nodule detection. The input of the network is the two-channel (image data and mask) 3D 128 × 128 × 128 tensor. Two loss functions (shown in Eq. (3), CEL and CRL) are trained jointly.

## 4. EXPERIMENTS AND RESULTS

### 4. 1 Data Distribution

In this paper, we evaluate our method on the joint cohort of two clinical datasets: Consortium for Molecular and Cellular Characterization of Screen-Detected Lesions (MCL) and Vanderbilt Lung Screening Program (VLSP). In total, we have

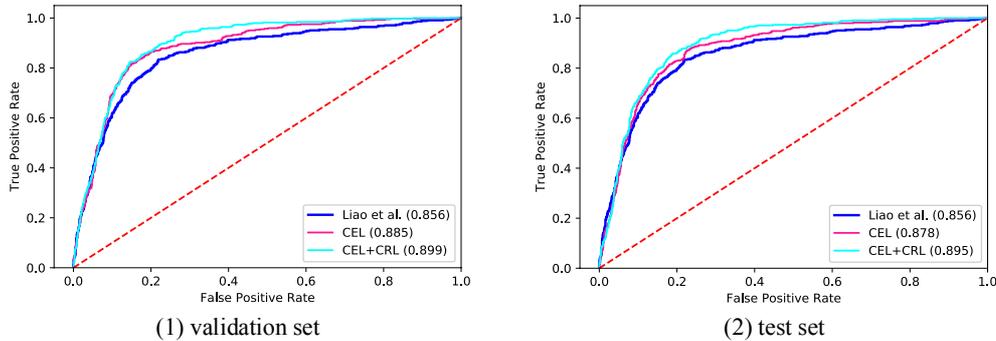

(1) validation set  (2) test set

Figure 4. The ROC curve of the different methods. (1) the performance on validation set reflect the fitting ability of our method, and (2) the performance on test set indicates the generalization ability of models.

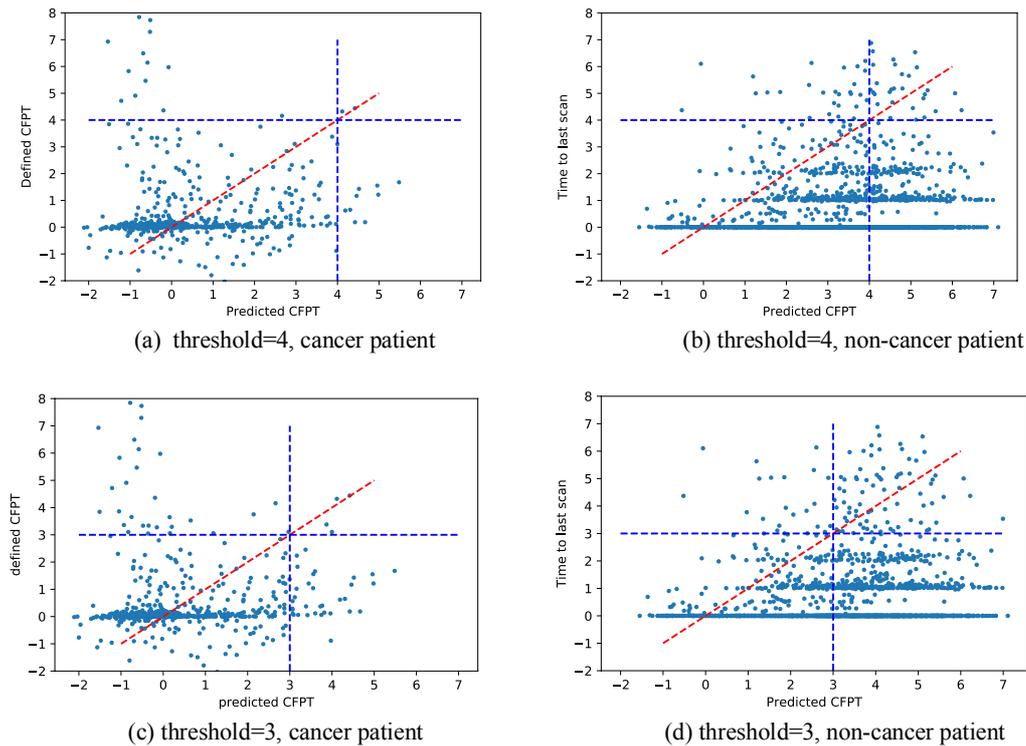

(a) threshold=4, cancer patient  (b) threshold=4, non-cancer patient

(c) threshold=3, cancer patient  (d) threshold=3, non-cancer patient

Figure 5. Analyses of CFPT prediction. (a) is predicting results of the scans from the cancer patients and threshold = 4. (b) is predicting results of the scans from the non-cancer patients and threshold = 4. (c) is predicting results of the scans from the cancer patients and threshold = 3. (d) is predicting results of the scans from the non-cancer patients and threshold = 3. The ratios of each region are shown in Table 1.

Table 1. Ratios (%) of different regions in Figure 5

|  | Region 1 | Region 2 | Region 3 | Region 4 |
|---|---|---|---|---|
| Figure 5(a) | 0.3 | 2.6 | 95.5 | 1.5 |
| Figure 5(b) | 1.7 | 1.6 | 59.5 | 37.1 |
| Figure 5(c) | 0.6 | 5.3 | 88.3 | 5.7 |
| Figure 5(d) | 4.5 | 1.3 | 34.4 | 59.7 |

Table 2. Classification performances with different thresholds (%)

| Threshold (year) | 1 | 2 | 3 | 4 | 5 |
|---|---|---|---|---|---|
| Recall | 60.07 | 76.76 | 88.30 | 95.52 | 97.93 |
| Non-cancer beyond threshold | 90.62 | 80.66 | 64.44 | 38.80 | 15.65 |

*The Recall is the region 3 percent of cancer-patient plot. The "Non-cancer beyond threshold" is the sum of region 1 and region 4 of non-cancer patient plot.

2287 scans from 1433 de-identified subjects, including 445 positive (cancer) scans and 372 final diagnosed cancer patients. All data are under internal review board supervision.

We randomly equal separate the whole dataset as 5 folds and apply five-fold cross-validation in the experiments. In each fold, 20% data (i.e., one-fold) are held out from training as the test set, and the remainder of the data is split as 3:1 for training and validation sets. Each fold is separated on the patient level, which the CT scans from the same patient will not distributed at different folds. The termination epoch is chosen based on the performance of the validation set, and directly applied on the test set. All the samples are included the test set once and only once, and final AUC are computed with the cancer probability of all test samples from the five folds (Figure 4).

### 4.2 Training details and Hyperparameters

Our experiments are implemented with Pytorch 1.0 and Python 3.6 using GTX TiTan X hardware. The max epoch number is 120. The Adam optimizer is chosen for training and the learning rate is initialized with 4e-5 and is multiplied by 0.4 at $40^{th}$, $60^{th}$ and $80^{th}$ epochs. The weight decay is set to 0.01 to avoid overfitting. The $\lambda$ and $\varepsilon$ in the loss function are set to 0.5 and 1, respectively. The best validation epoch is selected by the minimum validation loss.

### 4.3 Diagnosis Task

The receiver operating characteristic (ROC) curve shows the performance of a classification model. The area under the ROC curve (AUC) is used as the evaluation criteria of classification model. The Liao et al. [4] (Kaggle DSB2017 winner method) are included for comparison (directly apply the open source trained model). We show the AUC curve of validation set and test set. Note Figure 4 combines the all samples used in the five-fold testing cohorts such that no sample was used in training in a model for which the sample was used in testing. The improvement of the joint learning of CRL and CEL on single learning of CEL is significant (McNemar test, $p < 0.01$).

### 4.4 Prognosis Task

Figure 5 shows the predicted CFPT v.s. time to last scan (non-cancer patient) and predicted CFPT v.s. defined CFPT (cancer patient), which includes all the test sets from 5-fold cross-validation. The term "threshold" represents the value chosen on x- and y-axis to split the figures (Figure 5). Four regions are obtained with the threshold (blue line in the Figure 5, region 1: upper right, termed as CFPT-Conservation-Under; region 2: upper left, termed as CFPT-Conservation-Over, region 3: lower right, termed as CFPT-Optimistic-Over, region 4: lower right, termed as CFPT-Optimistic-Under).

From a machine learning perspective, since we deal with the censored data so the predicted CFPT might be biased by cancer patients. Thus, it is reasonable that the mean of the CFPT for non-cancer subjects is around 3.5 years, which is lower than the mean cancer free time 4.5 years that computed from the reported cancer free days of the over 50,000 NLST [2] subjects that not diagnosed as cancer. In the other hand, the CRL will not punish the non-cancer sampled with predicted

CFPT larger than defined CFPT + ε, and the mean predicted CFPT of non-cancer patients is clear higher that of cancer patients.

As shown Figure 5 and Table 1, the region 3 with 95.5% scans when the threshold set to 4, indicating 95.5% of within 4 years cancer patients are diagnosed with cancer. At the same time, about 40% non-cancer patient with the predicted year beyond the threshold 4 years, which indicates those scans with less 5% probability to have a lung cancer, and might be lower priority for screening.

Compared with certain year (e.g., 1 year, 2 years as in [5]) lung cancer prediction, our model can estimate results of any time as threshold without re-training the model. As shown in Table 2, five thresholds are select to illustrate the cancer / no-cancer prediction performance of following certain year.

## 5. CONCLUSION

In this paper, we propose a novel Censored Regression Loss (CRL) function to predict cancer-free progression time (CFPT). The proposed CRL has potential to add value to clinical interpretation of low dose lung screening CT. First, with the proposed approach, every scan is classified as cancer or non-cancer. The result is more accurate than using a single task model as the joint loss of CRL and cross-entropy loss bring significant improvement on classification performance. Second, for those patients with very large predicted CFPT, follow-up screening or biopsy could be suggested later based on our analyses (like non-cancer patient example in Figure 1). For example, if a patient has a predicted CFPT of more than five years, the risk of developing cancer is less than 4.5% one year later (see Table 1). Third, for patient with a low/ intermediate predicted CFPT, there could be increased emphasis on regular screening. One limitation of the proposed approach is the precision of CFPT given the current dataset. It would be interesting to extend learning to more of the NLST and other available low dose screening CT datasets. From a technical perspective, the loss functions are straightforward, but the generalizing across sites and with imbalanced data requires care.

## 6. ACKNOWLEDGEMENT

This research was supported by NSF CAREER 1452485 and NIH R01 EB017230. This study was supported in part by a UO1 CA196405 to Massion. This study was in part using the resources of the Advanced Computing Center for Research and Education (ACCRE) at Vanderbilt University, Nashville, TN. This project was supported in part by the National Center for Research Resources, Grant UL1 RR024975-01, and is now at the National Center for Advancing Translational Sciences, Grant 2 UL1 TR000445-06. We gratefully acknowledge the support of NVIDIA Corporation with the donation of the Titan X Pascal GPU used for this research. The imaging dataset(s) used for the analysis described were obtained from ImageVU, a research resource supported by the VICTR CTSA award (ULTR000445 from NCATS/NIH), Vanderbilt University Medical Center institutional funding and Patient-Centered Outcomes Research Institute (PCORI; contract CDRN-1306-04869). This research was also supported by SPORE in Lung grant (P50 CA058187), University of Colorado SPORE program, and the Vanderbilt-Ingram Cancer Center.

## 7. REFERNCES

[1] Miller, K.D., Siegel, R.L., Lin, C.C., Mariotto, A.B., Kramer, J.L., Rowland, J.H., Stein, K.D., Alteri, R., Jemal, A., "Cancer treatment and survivorship statistics, 2016." CA. Cancer J. Clin. 66, 271–289 (2016).
[2] N.L.S.T.R.T.J., "The national lung screening trial: Overview and study design." Radiology 258, 243–253 (2011).
[3] Tammemägi, M.C., Katki, H.A., Hocking, W.G., Church, T.R., Caporaso, N., Kvale, P.A., Chaturvedi, A.K., Silvestri, G.A., Riley, T.L., Commins, J., Berg, C.D.: "Selection criteria for lung-cancer screening." N. Engl. J. Med. 368, 728–736 (2013).
[4] Liao, F., Liang, M., Li, Z., Hu, X., Song, S.: "Evaluate the Malignancy of Pulmonary Nodules Using the 3-D Deep Leaky Noisy-or Network." IEEE Trans. Neural Networks Learn. Syst. , 1-12 (2019).
[5] Ardila, D., Kiraly, A.P., Bharadwaj, S., Choi, B., Reicher, J.J., Peng, L., Tse, D., Etemadi, M., Ye, W., Corrado, G., Naidich, D.P., Shetty, S.: "End-to-end lung cancer screening with three-dimensional deep learning on low-dose chest computed tomography," Nature Medicine 25(6), 954 (2019).
[6] Wang, J., Gao, R., Huo, Y., Bao, S., Xiong, Y., Antic, S.L., Osterman, T.J., Massion, P.P., Landman, B.A.: "Lung cancer detection using co-learning from chest CT images and clinical demographics." SPIE Medical Imaging 2019: Image Processing 10949, 109491G (2019).
[7] Gao, R., Huo, Y., Bao, S., Tang, Y., Antic, S.L., Epstein, E.S., Balar, A.B., Deppen, S., Paulson, A.B., Sandler,

|       | K.L., Massion, P.P., Landman, B.A.: "Distanced LSTM: Time-Distanced Gates in Long Short-Term Memory Models for Lung Cancer Detection," Proceedings of Machine Learning in Medical Imaging 11861, 310 (2019) |
| --- | --- |
| [8]  | Liu, L., Dou, Q., Chen, H., Olatunji, I.E., Qin, J., Heng, P.A.: "Mtmr-net: Multi-task deep learning with margin ranking loss for lung nodule analysis," Deep Learning in Medical Image Analysis and Multimodal Learning for Clinical Decision Support, 74-82 (2018). |
| [9]  | Tabibu, S., Vinod, P.K., Jawahar, C. V.: "Pan-Renal Cell Carcinoma classification and survival prediction from histopathology images using deep learning," Sci. Rep. 9, (2019). |
| [10] | Kaplan, E.L., Meier, P.: "Nonparametric Estimation from Incomplete Observations," J. Am. Stat. Assoc. 53, 457–481 (1958) |
| [11] | Schroff, F., Kalenichenko, D., Philbin, J.: "FaceNet: A unified embedding for face recognition and clustering," Proceedings of the IEEE Computer Society Conference on Computer Vision and Pattern Recognition , 815–823 (2015). |
| [12] | Liu, W., Wen, Y., Yu, Z., Li, M., Raj, B., Song, L.: "SphereFace: Deep hypersphere embedding for face recognition," Proceedings of IEEE Conference on Computer Vision and Pattern Recognition , 6738–6746 (2017). |
| [13] | Gao, R., Yang, F., Yang, W., Liao, Q. "Margin Loss: Making Faces More Separable," IEEE Signal Process. Lett. 25, 308-312 (2018). |